\pdfoutput=1

\documentclass[11pt]{article}
\usepackage{times,latexsym}
\usepackage{url}
\usepackage[T1]{fontenc}
\usepackage{url}
\usepackage{amsmath}
\usepackage{amssymb}
\usepackage{multirow}
\usepackage{float}
\usepackage{color,soul}
\usepackage{tikz}
\usepackage{subcaption}
\usepackage{graphicx}
\usepackage{array}
\usepackage{booktabs}
\usepackage{mathtools}
\usepackage{bbm}

\DeclareMathOperator{\pmi}{PMI}

\usepackage{acl}

\usepackage{times}
\usepackage{latexsym}

\usepackage[T1]{fontenc}

\usepackage[utf8]{inputenc}

\usepackage{microtype}

\usepackage{inconsolata}

\usepackage{graphicx}

\title{{\sc ConTestS}: a Framework for Consistency Testing of \\ Span Probabilities in Language Models}

\author{Eitan Wagner$^\dagger$\quad Yuli Slavutsky$^\ddagger$\quad Omri Abend$^\dagger$ \\
         $^\dagger$ School of Computer Science and Engineering \\ \quad
         $^\ddagger$ Department of Statistics and Data Science\\
         Hebrew University of Jerusalem\\ \texttt{\{first\_name\}.\{last\_name\}@mail.huji.ac.il}}

\begin{document}
\maketitle
\begin{abstract}
Although language model scores are often treated as probabilities, their reliability as probability estimators has mainly been studied through calibration, overlooking other aspects. In particular, it is unclear whether language models produce the same value for different ways of assigning joint probabilities to word spans. Our work introduces a novel framework,  ConTestS (Consistency Testing over Spans), involving statistical tests to assess score consistency across interchangeable completion and conditioning orders. We conduct experiments on post-release real and synthetic data to eliminate training effects. Our findings reveal that both Masked Language Models (MLMs) and autoregressive models exhibit inconsistent predictions, with autoregressive models showing larger discrepancies. Larger MLMs tend to produce more consistent predictions, while autoregressive models show the opposite trend. Moreover, for both model types, prediction entropies offer insights into the true word span likelihood and therefore can aid in selecting optimal decoding strategies. The inconsistencies revealed by our analysis, as well their connection to prediction entropies and differences between model types, can serve as useful guides for future research on addressing these limitations.\footnote{Code is provided at \url{https://github.com/eitanwagner/contests}}
\end{abstract}

\section{Introduction}
Pretrained Large Language Models (LLMs) emerged as high-performance predictors for diverse tasks \cite{brown2020language}. Tuning based on instructions and human feedback has further advanced their performance \cite{wei2022finetuned, ouyang2022training}.
In various applications, the model's success often depends solely on assigning high scores to the correct options. Yet, their interpretation as probabilities is beneficial in several applications, such as their treatment as confidence indicators for detecting hallucinations \cite{manakul-etal-2023-selfcheckgpt, ren2023robots} and robust ranking in sequence generation tasks \cite{zhao2022calibrating}.

Indeed, LLMs are commonly trained using the cross-entropy objective, which due to Gibbs' inequality is minimized by the true distribution. However, this global minima is hard to achieve \cite{chang-mccallum-2022-softmax, zhu2023calibration}
, raising the question of whether the model's outputs can still be interpreted as estimated probabilities.

\begin{figure}[t]
\centering
\includegraphics[scale=0.148]{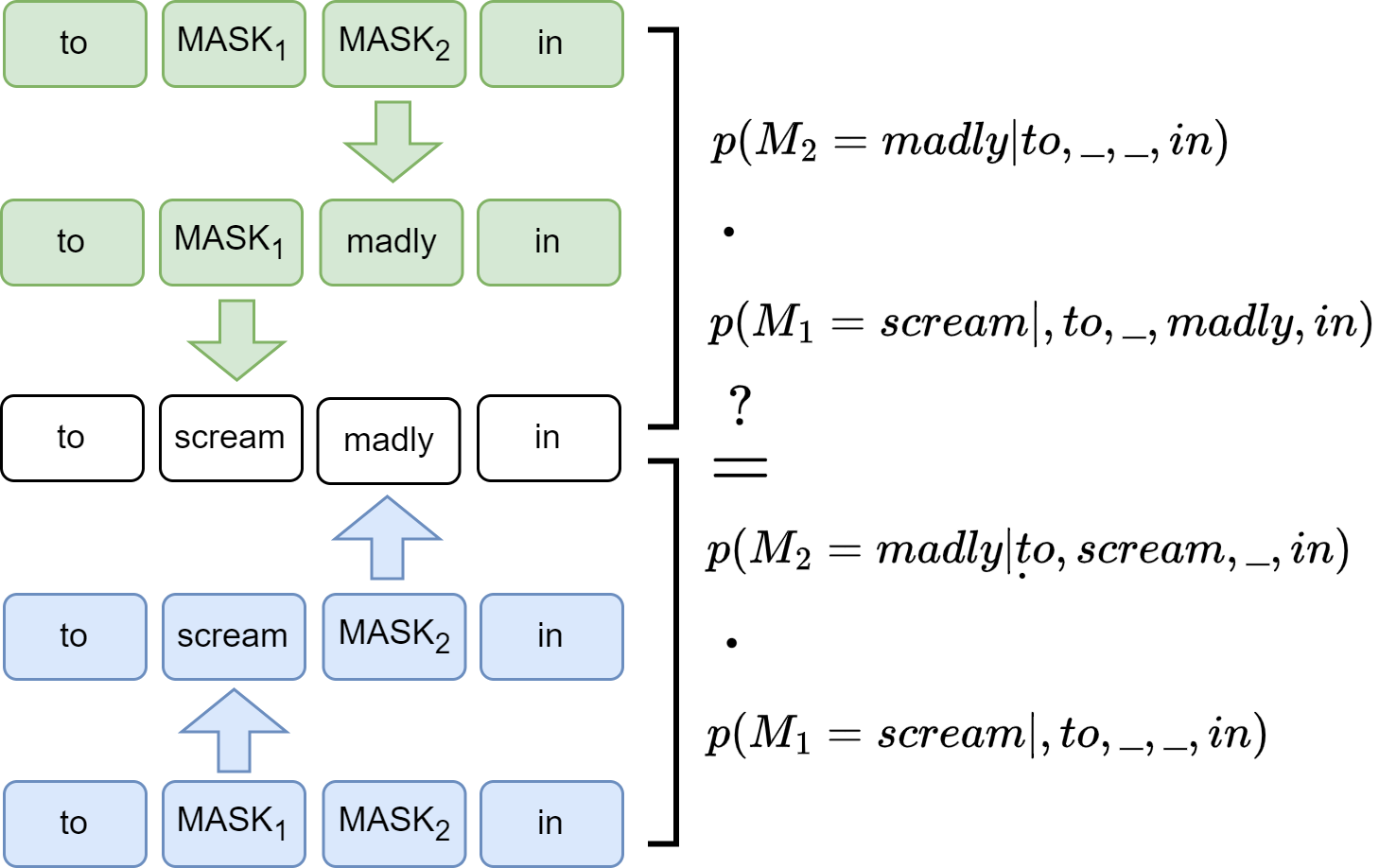}
\caption{Experimental Design - Joint Prediction Estimation with Masked Language Modeling.
The middle white row displays the original unmasked tokens. Below, in blue, the joint probability is calculated by first estimating the probability of the correct token in MASK$_1$ and then of MASK$_2$ (after revealing the correct token in MASK$_1$). In the top rows, in green, the calculation is in the reversed order -- first estimating the probability of the correct token in MASK$_2$ and then in MASK$_1$ (after revealing MASK$_2$).}
\label{fig:figure1}
\end{figure}

While calibration of produced scores has been extensively studied \cite{zhao2022calibrating, shen2024thermometer}, multiple other aspects remain unexplored. Measuring calibration for string density estimation requires a ground-truth measure for the real distribution, which is challenging to obtain (see \S \ref{calibration}). Therefore, alternative methods to validate the assumption that produced scores correspond to estimated probabilities, are needed.

For LLM scores to be interpreted as probabilities, consistency across estimation methods is essential.
However, regardless of probabilistic interpretation, detecting and understanding inconsistencies among estimation methods is crucial, especially in the identification of preferable estimation methods. 

Considering completion of word spans, consistency implies that filling masks in different orders (first filling one word and then the other, or vice versa) produces the same joint probability (see Figure \ref{fig:figure1}). In this work, we investigate whether this requirement is fulfilled.

Various factorizations of a joint distribution into conditional probabilities are possible, for instance by applying the chain rule from left-to-right or right-to-left. However, an unrestricted set of conditional probabilities does not guarantee a unique joint probability (see example in Appendix \ref{appendix: A1}). Although restricting the conditional probabilities, such as by disallowing cycles in Bayesian Networks, can ensure a unique joint probability, it is not necessary (see example in Appendix \ref{appendix: A2}).

\emph{Masked language modeling} (MLM) training lacks mechanisms to ensure that a set of conditional probabilities will form a unique joint distribution. However, since language modeling is based on the assumption that a distribution over strings generates the samples, estimated conditional distributions are expected to align with joint conditioning.

We investigate the consistency of both MLM and autoregressive models (which include decoder-only and encoder-decoder models), considering that MLM can function as a missing token classification task or a generative task with a specific instruction prompt (see \S \ref{sec: prob-mlm}). When treating autoregressive models, we account for task comprehension as a contributing factor.

We introduce a novel framework that employs statistical tests -- {\sc ConTestS}, for Consistency Testing over Spans, to analyze discrepancies between different estimation methods, and their behaviors across various model types.

Our findings show that all examined LLMs fail to produce consistent probabilities for joint span modeling. However, we observe notable distinctions among model types and sizes: autoregressive models show increasing discrepancies as model parameters increase, while MLMs tend to provide more consistent predictions, with larger models offering further improvements. Additionally, we show that prediction entropies are indicative of the true likelihood for both model types, suggesting their usefulness in selecting optimal decoding strategies. 

\section{Preliminaries and Notation}
The task of single-mask probabilistic masked language modeling is to estimate the probability of a masked location, given the rest of the sequence. 
For a model $M$ and  a sequence of tokens $\mathbf{x} = (x_1, x_2 ... x_n)$, we denote the estimation by 
\begin{multline*}
    P_M(x_i=w_i|x_1=w_1,\ldots ,x_{i-1} = w_{i-1}, \\ x_{i+1}=w_{i+1},\ldots,x_n=w_n).
\end{multline*}

When multiple masks are considered, that is, when in addition to $i$, the positions $j_1,..,j_k$ are masked as well, the predicted distribution for the $i$-th position given all unmasked ones is given by
\begin{equation*}
    P_M(x_i=w_i| x_j=w_j \forall j \notin \{i,j_1,\ldots,j_k\}),
\end{equation*}
and the joint distribution of two masked positions $i, j$, given unmasked ones, is denoted by
\begin{equation*}
   P_M(x_i=w_i, x_j=w_j| x_k=w_k \forall k \notin \{i,j\}).
\end{equation*}

Masked language modeling can be performed with autoregressive models, predicting a span given an appropriate instruction prompt. These models predict $P_M(x_{i}=s|\mathbf{x}_{-i})$, where $s$ is a span of arbitrary length (i.e., $len(s) \geq 1$).\footnote{We emphasize that the mask is a single token while the predicted span can be longer.} The predicted length is determined by the prediction of an end-of-sentence (EOS) token. As autoregressive models are trained to estimate the probability of completions, the probability of the token followed by EOS should match the MLM probability.\footnote{Like MLMs, autoregressive models are trained using cross-entropy loss, which is minimized by the true distribution over possible completions. Instruction tuning introduces additional objectives, but the model is regularized to stay close to its pretrained state, suggesting that consistency of the scores is a reasonable expectation in these models as well.}

Although we define the masked language modeling task regardless of the prediction model, following common convention, we will use the simple term MLM for models that were pretrained with the MLM objective. We will explicitly mention the use of autoregressive models.

For simplicity, we focus on the estimation of joint probabilities of two tokens.
To neutralize the effect of word distances, we analyze probabilities of adjacent tokens $P(x_i, x_{i+1})$, which can be expressed in two forms:
\begin{align} \label{eq:P1}
P^{(\mathbf{x})}_{i,i+1}(x_{i},x_{i+1})\coloneqq 
& P(x_{i}|\mathbf{x}_{-x_i,-x_{i+1}})\\ \nonumber
 & \cdot P(x_{i+1}|\mathbf{x}_{-x_{i+1}})\\
P^{(\mathbf{x})}_{i+1,i}(x_{i},x_{i+1})\coloneqq & \label{eq:P2}P(x_{i+1}|\mathbf{x}_{-x_{i},-x_{i+1}})\\ \nonumber
 & \cdot P(x_{i}|\mathbf{x}_{-x_{i}}).
\end{align}
When the identity of $x_{i}$ and $x_{i+1}$ is implicitly clear we denote these expressions simply as  $P_{i,i+1}$ and $P_{i+1,i}$.
Although for the true distribution we have that $P_{i,i+1} = P_{i+1,i}$, this may not hold for estimated probabilities since each direction involves two inference steps, each with slightly different inputs.
\footnote{We note that our framework can easily be extended to token sequences of arbitrary length and not necessarily adjacent. For a sequence of $n$ tokens, we can mask them all and estimate the joint probability by sequential estimation of conditional probabilities. A well-defined probabilistic model will give the same probability regardless of the order, yielding a total of $n!$ estimations.}

When analyzing discrepancies, we at times consider the 
\emph{pointwise mutual information} (PMI) between two random variables at $x_1\in \mathcal{X}_1, x_2 \in \mathcal{X}_2$:
\begin{equation} \label{eq: pmi}
    \pmi(x_1, x_2) = \log\frac{P(x_1, x_2)}{P(x_1)P(x_2)}
\end{equation}
and for the joint distribution, we have: 
\begin{small}
\begin{multline*}
P(x_1)\cdot P(x_2|x_1)=P(x_1,x_2)=P(x_2)\cdot P(x_1|x_2) \\ 
\Leftrightarrow \log\frac{P(x_1|x_2)}{P(x_1)} = \log\frac{P(x_1,x_2)}{P(x_1)P(x_2)} = \log\frac{P(x_2|x_1)}{P(x_2)}.
\end{multline*}
\end{small}

\section{Previous Work}

\paragraph{Consistency.}\label{consistency}

Few works have directly addressed the assessment of consistency. Among those that have, many focused on testing whether outputs adhere to predefined constraints.
\citet{li-etal-2019-logic} and \citet{ribeiro-etal-2019-red} demonstrated violations of logical constraints in question-answering. \citet{elazar-etal-2021-measuring} evaluated the internal knowledge consistency of language models (LMs) comparing outputs for paraphrases of semantically identical questions. \citet{pezeshkpour2023large} demonstrated sensitivity to answer ordering in multiple-choice questions. \citet{qiu2023large} illustrated inconsistent temporal reasoning in LLMs across various time-related tasks.

In our work, we focus on the consistency of probabilities rather than outputs. While identical probabilities imply identical outputs, the reverse is not necessarily true in language modeling, making our approach more sensitive to inconsistencies.

\paragraph{Calibration.}  \label{calibration}
A common approach to assessing the quality of predicted probabilities is through calibration, which evaluates how well predicted probability scores align with membership probabilities in some reference data. In \textit{fully calibrated}  
multi-class classifiers, calibration is considered for every class in every prediction. However, evaluating calibration, even in binned probabilities, becomes challenging with a large number of classes, making meaningful binning for every class with representative data difficult. To address this, many studies opt for \textit{top-class} calibration \cite{guo2017calibration}, which focuses solely on calibrating the predicted class.

Although top-class calibration is sufficient to assess the confidence of the prediction, and therefore is frequently used \cite{jiang2012calibrating, guo2017calibration},  full calibration is an essential requirement for a model to be used as a density estimator in multi-class classification, in structured predictions such as sequence predictions in autoregressive text generation, and complex probabilistic generative models with textual components.

While measuring full calibration directly is challenging, our approach, which compares the consistency of assigned probabilities to the same expression using different methods, offers alternative means to identify uncalibrated models -- inconsistent estimations across different methods imply that at least one of them is miscalibrated.

Many works measured the calibration of neural models \cite{guo2017calibration, wang2021rethinking}, generally finding that neural models are poorly calibrated \cite{chen2023close}.  
In LMs, most prior work on calibration has focused on downstream tasks, such as classification and question answering \cite{desai-durrett-2020-calibration, dan-roth-2021-effects-transformer}. Studies that specifically addressed language modeling typically restricted their evaluations to top predictions \cite{zhu2023calibration}, top-prediction sets \cite{ravfogel-etal-2023-conformal}, or aggregate measures like entropy rates \cite{braverman2019calibration}. These evaluations have primarily examined autoregressive models, consistently finding them to be miscalibrated. While some research on masked language models (MLMs) has suggested that they tend to be relatively well-calibrated \cite{he2023preserving}, 
to the best of our knowledge, full-distribution calibration in language modeling was never addressed. 

\paragraph{Language Models as Density Estimators.} 
Several studies have interpreted language models as density estimators and explored their probabilistic properties from a theoretical standpoint.
\citet{hahn-2020-theoretical} proved that some cases cannot be efficiently modeled by Transformers. \citet{du-etal-2023-measure} showed that the requirement that infinite length strings will have zero probability might not be held for all models. However, they defined a theoretical notion of tightness that is satisfied by most common models and guarantees the requirement. \citet{wang-cho-2019-bert} showed that MLMs can be interpreted as Markov Random Fields, thus providing probabilities for entire sentences.
In contrast, \citet{yang2018breaking} showed theoretically that decoding based on softmax yields low-rank approximations that are inadequate for capturing the complexity of language distribution. Additionally, \citet{chang-mccallum-2022-softmax} presented findings indicating that decoding based on a single embedding vector cannot generate arbitrarily rich distributions. 

Research investigating sampling-based text generation includes the work of \citet{zhang-etal-2021-trading} that showed that sampling from an LM distribution results in low-quality text, and that the use of temperature scaling provides a tradeoff between quality and diversity. Similarly, \citet{Holtzman2020The} proposed nucleus sampling, avoiding the tail of the distribution, and \citet{meister-etal-2023-locally} presented a sampling scheme based on the expected entropy of natural language.

Several studies found conflicts between calibration and zero-shot capabilities. \citet{zhu2023calibration} showed that instruction tuning significantly hurts calibration.
\citet{kalai2023calibrated} proved that strict calibration, with respect to the training data, must lead to hallucinations. 
\citet{lee-etal-2020-discrepancy} showed a discrepancy between the cross-entropy loss, used for language modeling, and task-specific losses. 

Other work disregarded the probabilistic nature altogether, filling masks with spans from a reference document \cite{min-etal-2023-nonparametric}. This comes with the price of losing qualities of probabilistic estimation. 

\section{Desired Properties of a Consistency Testing Framework} \label{sec:requirements}

Our goal is to evaluate whether LLMs maintain consistency across different estimation orders when calculating the joint probability of a word span.
To ensure the robustness and reliability of this evaluation, it must meet the following requirements.

\paragraph{Versatility.} \label{sec: prob-mlm}
For the designed framework to apply to various models, it should address both MLMs and autoregressive models. It should account for task comprehension in autoregressive models, which are not specifically built for filling masks.

\paragraph{Significance of discrepancies.}
Minor variations in estimated probabilities may arise due to numerical issues. Additionally, discrepancies that are symmetrically distributed around zero, lacking a clear bias, may not have significant implications. Therefore, the analysis should prioritize statistically significant discrepancies.

\paragraph{Nullifying the impact of exposure in training.}
To prevent bias from analyzing data examples used in model training, evaluations should incorporate natural datasets that were not part of the model's training set.

\paragraph{Explainability.}
To be effective in both identifying and addressing inconsistencies, the framework should offer insights into contributing factors of found inconsistencies, such as model types, sizes (in parameters), and training data sizes. It should be able to isolate the effect of each factor, keeping the contribution of others fixed.

\section{Consistency Testing Framework}

We present the {\sc ConTestS}
framework, designed to meet the requirements outlined in \S \ref{sec:requirements}. Here, we detail how each requirement is addressed.

\subsection{Task Comprehension in Autoregressive Models}

Autoregressive language models are typically trained for next-token prediction and not directly for Masked language modeling. However, the MLM task was previously formulated as a conditional case of autoregressive language modeling -- the T5 model \cite{raffel2020exploring} was pretrained to generate text spans when given sentinel tokens in the input, and \citet{bavarian2022efficient} proposed training decoder-only models with a prompt for text infilling.  This formulation allows us to derive the distribution of the missing span by estimating the sequence of next-token probabilities.

Since autoregressive masked language modeling depends on task comprehension, we examine whether autoregressive models rank the true word sequence similarly to MLMs.
Additionally, since these models allow for predictions of multiple tokens even when asked to fill one only, we analyze the scores assigned to EOS as the second token. A high probability for EOS as the second token is a positive indicator for understanding the task.

\subsection{Testing Discrepancy Significance}

Since large language models usually provide small probabilities, and due to the connection between the expressions for the joint  to the expression for PMI (see Equation \ref{eq: pmi}), here we examine the consistency of a given model by the discrepancy between estimations of the two expressions in log scale
\begin{align}
d_{i,i+1}(x_{i},x_{i+1})\coloneqq & \log P_{i,i+1}(x_{i},x_{i+1})\\
 & -\log P_{i+1,i}(x_{i},x_{i+1}). \nonumber
\end{align}

For each text $\mathbf{x}^{(j)}$ and a  pair of consecutive tokens\footnote{When the identity of $i$ is immaterial, we simply use the notation $d^{(j)}$.} $x_{i},x_{i+1}$ 
we compute $d^{(j)}_{i,i+1}(x_{i},x_{i+1})$.
As discrepancies $d^{(j)}$ are functions of the random variables $\mathbf{x}^{(j)}$, they follow an unknown distribution $f$.
For a perfectly calibrated model, $d^{(j)}=0$ for all $j$, indicating a singleton mass of $f$ at 0.
In practice, the distribution induced by a model $M$ is unknown and requires non-parametric treatment. Therefore, to test for the consistency of $M$, we employ the paired two-sided Wilcoxon rank test \cite{wilcoxon-45} to assess the null hypothesis that $f$ is symmetric around 0.
That is, 
\begin{align} \label{eq:hypothesis}
H_{0} & :f\text{ is symmetric around }\mu=0\\
H_{1} & :f\text{ is symmetric around }\mu\neq0.
\end{align}

To examine these hypotheses, the Wilcoxon test employs the test statistic $T=\sum_j \text{sgn}(d^{(j)})R^{(j)}$ where $R^{(j)}$ is the rank of $d^{(j)}$ (i.e., position in the sorted array) and $\text{sgn}(d^{(j)})=1$ if $d^{(j)}>0$, $-1$ if $d^{(j)}<0$ and 0 otherwise.
The reliance of the Wilcoxon test on ranks
carries the following implications: (1) it demonstrates robustness to extreme discrepancy values, and (2) the focus on ranks, rather than the discrepancy values themselves, increases the difficulty of rejecting the null hypothesis, rendering it a conservative test. 

In our analysis, we apply the Wilcoxon test to assess the significance of discrepancy means for multiple models across various text datasets. 
Given that testing a model across multiple datasets, and testing multiple models on the same dataset, increases the risk of type I error (the mistaken rejection of a true null hypothesis), we conduct a correction for multiple comparisons. We use the relatively conservative correction proposed by \citet{benjamini-yekutieli} since it applies to dependent tests.

\subsection{Data Gathering} \label{gathering}
To eliminate biases caused by exposure to data in training, in addition to benchmark and synthetic datasets, we constructed a new dataset by extracting \textbf{news articles} with topics ``WORLD'', ``NATION'', ``BUSINESS'', ``TECHNOLOGY'', ``ENTERTAINMENT'', ``SCIENCE'', ``SPORTS'', and ``HEALTH'' from Google-news\footnote{\url{https://news.google.com/}}. Extraction was conducted on dates after the models' training data cutoff.

\subsection{Testing for Contributing Factors}

To analyze differences between different model types, while isolating factors such as their different number of parameters or training volume, we conducted a linear regression analysis.
This approach is chosen because, in linear regression, a coefficient represents the change in the dependent variable associated with a one-unit change in the independent variable, with all other variables held fixed.
 
For all considered models $1, \dots, K$, let $d^{(j)}_k$ represent  discrepancy value computed for the $k$-th model and a text $\mathbf{x}^{(j)}$. for $1 \leq j \leq J$. We set the model type $T_k$ to 1 if the model is autoregressive, and set $T_k=0$ if it is an MLM. We investigate the influence of the model type, parameter size $S_k$  (in billions of parameters), and the size of the training set (in GB) $V_k$ on the variance of discrepancy
\begin{equation}
    \nu_k(d) \coloneqq \text{Var}(d^{(1)}_k, \dots, d^{(J)}_k).
\end{equation}
In this analysis, we consider the variance $\nu_k$ as the dependent variable, while the other parameters serve as explanatory variables:
\begin{align} \label{eq:regression}
\hat{\nu}_{k}= & \hat{\beta}_{0}+\hat{\beta}_{1}S_{k}+ \hat{\beta}_{2}V_{k}+\hat{\beta}_{3}T_{k} +\hat{\beta}_{4}S_{k} T_{k} 
\end{align}

Whether a model adequately captures the variability of the explained variable, is often measured by $0 \leq R^2 \leq 1$. When the model is indeed well fitted ($R^2$ close to 1), large and statistically significant coefficients indicate a substantial effect of the corresponding explanatory variables (in our case the effect of model size within specific types).

For an autoregressive model, the estimated variance is 
$$\hat{\nu}_{k}=(\hat{\beta}_{0}+ \hat{\beta}_{3})+\hat{\beta}_{2}V_k+(\hat{\beta}_{1}+\hat{\beta}_{4})S_{k},$$ whereas for an MLM it is
$\hat{\nu}_{k}=\hat{\beta}_{0}+\hat{\beta}_{1}S_{k}+\hat{\beta}_{2}V_k$. 
Consequently, the effect of the number of parameters in MLMs is incorporated in $\hat{\beta}_{1}$ and serves as the baseline effect, against which autoregressive models are estimated.

Note, that in contrast to discrepancy testing where the analysis is performed for a given model and the observations are texts in the dataset, here the models themselves are treated as observations.

\section{Experimental Design} 

\begin{figure*}[t]
  \centering
  \begin{subfigure}{0.73\textwidth}
  \includegraphics[width=\textwidth]{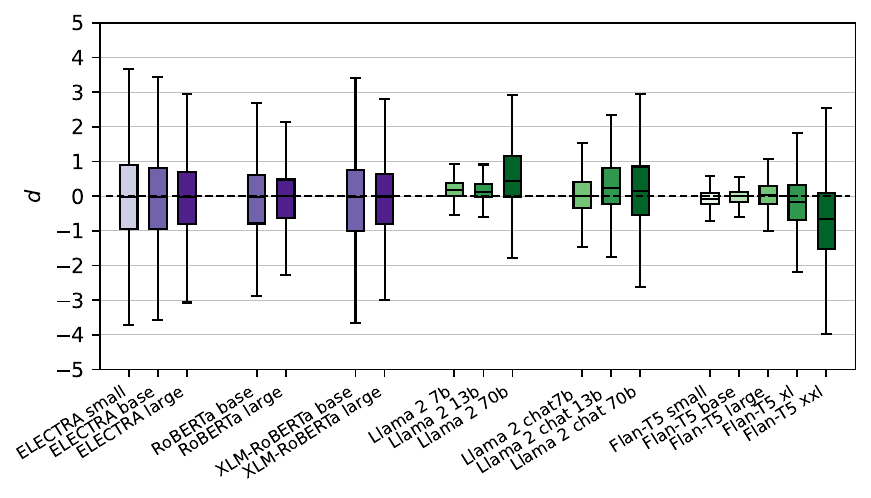}
    \caption{Wikitext}
    \label{fig:subfig_a}
  \end{subfigure}
    \begin{subfigure}{0.73\textwidth}
    \includegraphics[width=\textwidth]{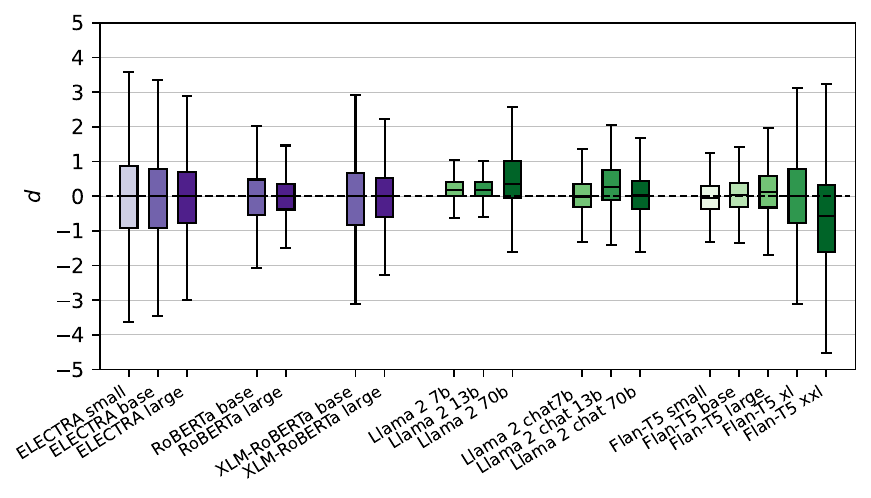}
    \caption{News}
    \label{fig:subfig_b}
  \end{subfigure}
  \vspace{2.mm}
    
  \caption{Discrepancy Results on the (a) Wikitext and (b) News datasets. Each model is represented by a boxplot displaying discrepancy values. MLMs appear in purple shades on the left of each figure and autoregressive models in green on the right. Color intensity indicates model sizes. Boxes show quartile values with median lines; whiskers extend to 1.5 IQR from quartiles. Outliers are omitted for clarity.}
  \label{fig:d_boxplots}
\end{figure*}

\subsection{Examined Models}
\paragraph{MLMs.}
We conducted experiments with the following MLMs, each in multiple sizes (parameters): RoBERTa \cite{liu2019roberta} base (125m parameters) and large (355m); XLM-RoBERTa \cite{conneau2020unsupervised} base (280m) and large (550m); we also used the generator component from ELECTRA \cite{clark2020electra} small (14m parameters in the generator) base (110m), and large (335m).

MLMs are typically trained with randomly masked tokens, including the possibility of adjacent masks \cite{devlin-etal-2019-bert}. Consequently, the probabilities used in our experiments align with the model’s training structure. For two adjacent masks, the predicted probabilities should represent the marginal distributions -- i.e., the likelihood of each token being correct, given the uncertainty of the adjacent token. When a token is masked independently of adjacent tokens, the prediction scores should reflect the probability conditioned on the surrounding context.

\paragraph{Autoregressive models.} We performed experiments using Flan-T5 \cite{chung2022scaling} small, base, large, xl, and xxl; {\sc Llama 2} and {\sc Llama 2-Chat} \cite{touvron2023llama} with 7b, 13b, and 70b parameters each. We used 4-bit quantization for all the autoregressive models, with nested quantization for the {\sc Llama 2} ones. These models suit our experiments as they include multiple sizes for the same architecture and settings.
We tested various prompts but did not notice notable differences. Prompts for the reported results are in Appendix \ref{appendix: A}.

We note that a natural option to test decoding order is to use T5's special tokens, putting them by order or in reverse. This method proved problematic, as T5 seems highly biased towards by-order completion, the format it was trained on. Reversing the order lowered the likelihood for all word pairs, with a stronger impact on high-probability pairs (e.g., common phrases), thus introducing a confounding reason for the discrepancy. 

\subsection{Data}

\paragraph{Natural Text.} We tested consistency over two datasets with texts from a natural source.
The first was \textbf{Wikitext-2}\footnote{\url{https://huggingface.co/datasets/wikitext/viewer/wikitext-2-raw-v1}} dataset, where we ignored punctuation, stop-words, and tokens that were not whole words. We used the train section, consisting of ${\approx} 37K$ articles. 
Since Wikitext was used during the training of the models, 
we constructed a new dataset as described in \S \ref{gathering} for four dates: 2.7.2023, 6.7.2023, 4.9.2023 and 18.9.2023, all well after the models' data cutoff. The texts were pre-processed in the same manner as in Wikitext. Altogether, the News dataset consists of ${\approx} 2000$ articles.

The exact set of word pairs evaluated in each dataset differed between models as different model types have different tokenizers. Model size (for a given type) does not affect tokenization. Additionally, due to computational limitations, a smaller set was used for the larger models.
In summary, for RoBERTa and ELECTRA-generator we had ${>} 200K$ word pairs in Wikitext and ${>} 85K$ in News, for XLM-RoBERTa ${\approx} 110K$ in Wikitext and ${\approx} 44K$ in News, and for Flan-T5 ${\approx} 175K$ in Wikitext and ${\approx} 55K$ in News. For {\sc Llama 2} (chat and non-chat) in Wikitext, we had ${\approx} 85K$ word pairs for the 7b and 13b versions and ${\approx} 13K$ for the 70b version, and in News, we had ${\approx} 11K$ for all sizes.

\paragraph{Synthetic Data.} We performed all experiments on synthetic data as well, in which the context is fixed for all samples. See Appendix \ref{appendix: B1} for details.

\section{Main Results}

\subsection{Consistency}

Here we provide an analysis of the distribution of discrepancy values $d^{(j)}$ for each examined model on real datasets. Analysis of the synthetic dataset is available in Appendix \ref{appendix: B2}.

The results of the experiments, summarized in Figure \ref{fig:d_boxplots}, show that on both Wikitext (\ref{fig:subfig_b}) and News (\ref{fig:subfig_a}) datasets, the distribution of discrepancies in MLMs is characterized by medians close to 0 and high variance, while autoregressive models exhibit discrepancies with medians further away from 0, but often lower variances.

The Wilcoxon test results on real datasets show that, except for Llama 2-chat-7b (p-value = 0.5997) and Flan-T5-xl (p-value = 1.0), both on the News dataset, all other models attain a corrected p-value smaller than 0.05 on both datasets.
Consequently, although MLMs exhibit median estimators only slightly different than 0, the null hypothesis (see equation \ref{eq:hypothesis}) of the distribution of discrepancies being centered around 0 is rejected for all these models. The highest corrected p-value among models where the null hypothesis is rejected is 0.0042, achieved by RoBERTa-large on the News dataset, indicating overall confidence levels above 99\%.

\begin{table} 
    \centering
    \resizebox{!}{0.6in}{
    \begin{tabular}{lcccc}
         \hline \hline
         & \multicolumn{2}{c}{Wikitext} & \multicolumn{2}{c}{News}\tabularnewline
          & Coeff & P-value & Coeff & P-value\\  
          \hline \hline
        Intercept  &0.7113	& \textbf{0.05} &1.4002	&\textbf{0.001} \tabularnewline
    Size & 2.7683	&\textbf{<0.001	}&72.2883	&\textbf{0.001} \tabularnewline
    Data size & -0.00002&	0.696	&-0.0001	&0.036\tabularnewline
    Type & 0.0196	&0.064	&0.0154	&0.126\tabularnewline
    I: Type -- Size & -2.4788	&0.228	&-3.0734	&0.131\tabularnewline
         \hline
    \end{tabular}
    }
    \caption{Analysis of Discrepancy Variance. Regression coefficients and corresponding p-values for each model are presented, with MLM models serving as the baseline. P-values below $\alpha=0.05$ shown in bold.}
    \label{tab:reg_results}
\end{table}

\subsection{Explaining Inconsistencies}

The results shown in figures \ref{fig:subfig_a}, \ref{fig:subfig_b} indicate that in real data, given a model type, MLMs exhibit smaller variance
as the number of parameters grows. Autoregressive models show an opposite trend. We performed a linear regression analysis (see equation \ref{eq:regression}) to test the significance of these trends, with the results summarized in table \ref{tab:reg_results}. 

The results on both datasets indicate with significance level $\alpha<0.001$ that the variance increases with the growth of the number of parameters. The size of the training set does not have a significant effect on the variance on the Wikitext dataset but influences the variance observed on the News dataset with a significance level of $\alpha=0.036$.

Both regression models capture the variability of $\nu_k$ values with $R^2$ values of $0.775$ and $0.794$ for the models fitted on the Wikitext and News datasets, respectively.

In appendix \ref{appendix: model_types} we provide a similar analysis, where instead of considering model types as autoregressive or MLMs, we consider their fine-grained model types (RoBERTa, ELECTRA, etc.).

\begin{figure*}[t]
\centering
\includegraphics[width=0.68\textwidth]{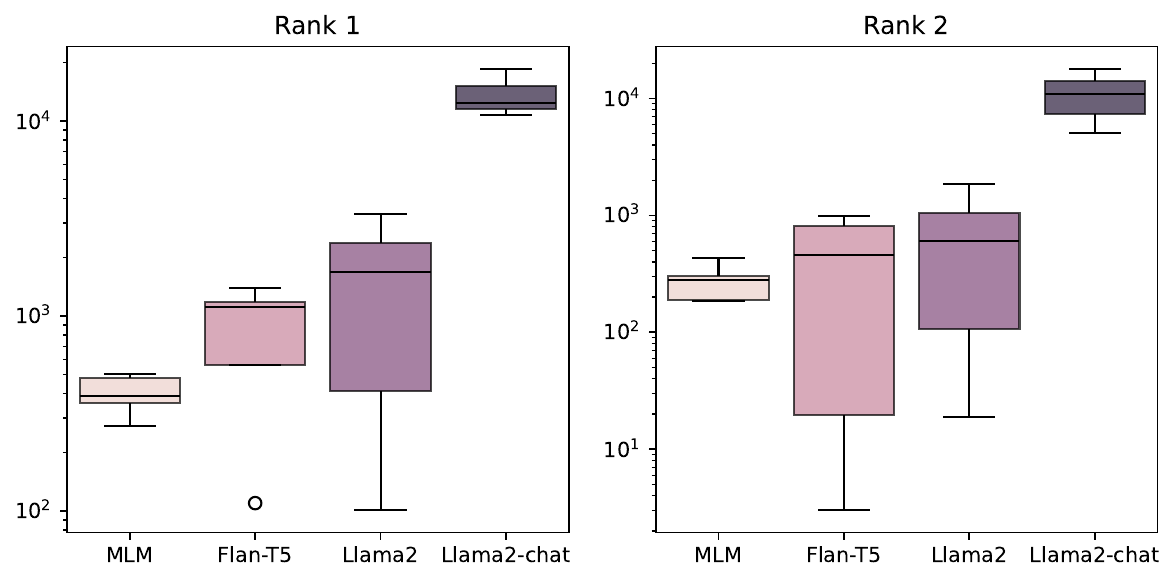}
\caption{Prediction Ranking for the examined Models. Rank 1 represents the ranks for the first prediction (two masks) and Rank 2 for the second (one mask). Results were obtained from a sample size of 200 on the News dataset. Ranks are in log scale. Lower ranks indicate more accurate predictions. Boxes show quartile values with median lines; whiskers extend to 1.5 IQR from quartiles. }
\label{fig:rank_boxplots}
\end{figure*}

\section{Task Comprehension in Autoregressive Models}

Our results show that autoregressive models tend to be less consistent, prompting the question of how much this inconsistency can be attributed to the more challenging task setting. To investigate this, we compare token ranks between MLMs and autoregressive models, where lower ranks indicate higher scores.
Ideally, both models should show similar rank distributions, indicating similar performance. Our analysis, shown in Figure \ref{fig:rank_boxplots}, indicates that while MLMs assign lower ranks to the true first token, Flan-T5, and {\sc Llama 2} models often assign even lower ranks to the second token, suggesting improved predictions. However, {\sc Llama 2-Chat} models exhibit significantly higher ranks, indicating poorer performance.\footnote{We emphasize that the comparison is not between the absolute ranks for the different prediction types, or between them and chance level, but rather between MLMs and autoregressive models relative to one another. We also note that we did not compare second-mask prediction where the decoding order is reversed between the two tokens.}

In Appendix \ref{appendix: C}, we provide a detailed analysis of probabilities assigned to an EOS token following the predicted missing word, as an additional measure of task comprehension. We find a positive correlation between models excelling in word rank predictions and those showing good task comprehension (indicated by lower EOS ranks). However, the poor performance of {\sc Llama 2-Chat} models cannot be solely attributed to a lack of task understanding, as they assign low ranks to EOS but high ranks to actual words. Across all model types, larger models in each category generally exhibit better task comprehension.

\section{Are there Preferable Decoding Orders?}

We showed that all examined models exhibit inconsistencies between different orders of estimation of joint probabilities of word spans. This raises the question of whether any of the examined completion orders yields higher scores for the true tokens.

\begin{figure}[t]
\centering
\includegraphics[width=0.49\textwidth]{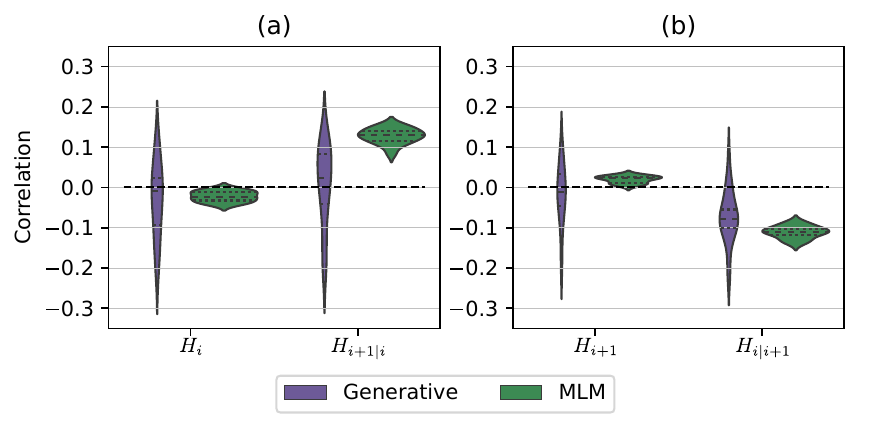}
\caption{
Entropy and Decoding Order: Correlations between four prediction entropies and the discrepancy $d_{i, i+1}$ are presented. (a) presents the two entropies associated with predicting the $i$-th token first. (b) illustrates the two entropies corresponding to predicting the $(i+1)$-th token first. For each entropy the distribution of correlations for autoregressive models is depicted in purple on the left, and for MLM models in green on the right. Dashed lines within each violin represent the first, second (i.e., median), and third quartiles, respectively.}
\label{fig:entropy}
\end{figure}

To address this question, we analyze the correlation between $d_{i,i+1}$ and the entropies of the estimated probabilities involved in the joint probability estimations, which we denote with $H_i, H_{i+1|i}, H_{i+1}, H_{i|i+1}$. A summary of the correlations is shown in Figure \ref{fig:entropy}. 

The analysis reveals that: (1) for entropies of predictions when two tokens are masked (i.e., $H_i, H_{i+1}$) the correlation with the discrepancy in the corresponding order ($d_{i,i+1}$, -$d_{i, i+1}$) is negative; (2) for one-mask entropies ($H_{i|i+1}, H_{i+1|i}$) the correlation is positive; (3) one-mask entropies exhibit stronger (in absolute value) correlations with discrepancy; and (4) correlations between entropies and discrepancy are stronger (in absolute terms) for MLMs compared to autoregressive models.

These results suggest that selecting the direction with higher entropy for one-mask prediction and lower entropy for two masks is likely to increase the likelihood of true tokens. In Appendix \ref{appendix: examples} we provide examples of the effect of decoding order.

\section{Discussion}

In this work, we investigated the probabilistic consistency of language models (LMs) and introduced a novel framework to quantify and explain discrepancies between equivalent estimation orders. 

Our findings indicate statistically significant inconsistencies among 16 out of 18 models on the News dataset and all 18 models on the Wikitext dataset. These results highlight significant differences between MLMs and autoregressive models, with the latter showing considerably larger inconsistencies and discrepancy variances.

The comparable discrepancy distributions between the News and Wikitext datasets suggest that exposure in training has little effect on consistency. In contrast, results from the synthetic dataset (see Appendix \ref{appendix: B2}) reveal a different pattern, with higher mean and variance in discrepancies, indicating reduced consistency on low likelihood data.

Analysis of the relationship between discrepancy variances and model size reveals significant trends only in Flan-T5 models, displaying a negative correlation. Across all model types, we found that larger models exhibit lower average prediction entropies. This suggests that artificially high consistency in autoregressive models may arise from high variance. However, this does not hold for MLMs, which show the opposite trend.

A positive correlation between real data likelihood and overall entropy is termed \textit{overconfidence}, while a negative correlation is called \textit{underconfidence} \cite{ravfogel-etal-2023-conformal}. Our experiments suggest that choosing a decoding direction with higher entropy for single-mask predictions and lower entropy for two masks is expected to yield higher estimated probabilities for the correct word pair. This suggests that overconfidence and underconfidence are influenced by the model type and the number of masked tokens, challenging the anticipated strong link between average likelihood and entropy.  Consequently,  this calls for cautious application of methods to address overconfidence and underconfidence, as models can exhibit both.

In conclusion, our investigation into the probabilistic consistency of language models has revealed significant inconsistencies across various model types and datasets. Our findings highlight the need for careful interpretation and application of the predicted scores. 

Prior research \cite{yang2018breaking} argued that modeling the ``distribution of language'' is a complex task. Our findings provide further support for this claim, showing that even high-performing models struggle to generate consistent estimations.

Robust ranking of structured predictions and similar applications often require consistency in joint probability estimation. However, our research shows that achieving high performance does not necessarily depend on consistency, and vice versa: a model can be perfectly consistent but make inaccurate predictions. Therefore, a combination of consistency testing with other performance metrics is essential for thorough evaluation.

Our analysis exposes inconsistencies and their links to prediction entropies and model-type disparities, offering valuable insights for future research to tackle these limitations effectively.

\section*{Limitations}
Our framework is designed for comparing any two estimations of a joint distribution. However, our experiments specifically target a setting with two adjacent tokens. While effective in revealing inconsistencies, exploring additional estimation orders with longer word spans in future research could uncover additional ones.

A limitation of the analysis of the dependency of discrepancy variances on model sizes lies in its treatment of each model as an observation, resulting in small sample sizes (7 MLMs, and 11 autoregressive models).

We also note that comparison between different model types is qualitative only, as they differ in their tokenization and other qualities (such as the scale of probabilities and entropies). In addition, due to the load of computation, in some cases, the sample sizes were relatively small.

\section*{Ethics Statement}
Wikitext-2 is released under license CC BY-SA 4.0.
The dataset of news articles was gathered from various new sites through Google News. The dataset was used for validation purposes only and will not be redistributed or used for training models. A list of URLs and access dates will be provided upon publication.

\section*{Acknowledgements}
The authors thank Gabriel Stanovsky, Amir Feder, and Noam Weis for their valuable insights.
This research was supported by grants from the Israeli Ministry of Science and Technology, the Israeli Council for Higher Education, and Israel Science Foundation (grant no. 2424/21).

\bibliography{anthology,custom}


\appendix

\section{Examples}
\subsection{Inconsistent Joint distribution from Conditionals} \label{appendix: A1}
Consider a (toy) example, where we have a two-word sentence $x_1, x_2$, over a yes-no alphabet $\{y, n\}$, and assume we have estimations $P(x_1=y|x_2=y)=0.9, P(x_2=y)=0.9$ and $P(x_2=y|x_1=y)=0.1, P(x_1=y)=0.1$. This leads to a contradiction, since 

\begin{small}
\vspace{-.2cm}
\begin{multline*}
0.81 = P(w_1=y|w_2=y) \cdot P(w_2=y) \\= P(w_1=y,w_2=y) \\= P(w_2=y|w_1=y) \cdot P(w_1=y) = 0.01.
\end{multline*}
\vspace{-.2cm}
\end{small}

\subsection{Consistent Joint distribution Without Structural Restrictions}  \label{appendix: A2}
As an example, consider a masked language model that is trained by maximum likelihood estimation with $(2n+1)$-grams. Since all conditionals are determined by the counts of the $(2n+1)$-gram samples, any set of resulting conditionals must comply with the $(2n+1)$-gram joint distribution.

\section{Prompts For Instruction Models}
\label{appendix: A}

The prompt we used is:
\begin{quote}
    You will be given a passage with one masked token that you should fill in. We denote this token by \%. The passage might also contain corrupted tokens denoted by @. You are not expected to fill in corrupted tokens - fill only the masked one. Your answer should include the filled-in token only with no extra explanations or context.
\end{quote}

For Flan-t5 and {\sc Llama 2} (non-chat version), the input format was:
\begin{quote}
    <prompt> \\
    Passage: <passage with masks> \\
    Answer:
\end{quote}

For {\sc Llama 2-Chat} the format was:
\begin{quote}
[INST] <<SYS>> \\
<prompt> \\
<</SYS>> \\ \\
Passage: <passage with masks> [/INST]
\end{quote}

\section{Explaining Inconsistencies With Model Types} \label{appendix: model_types}

In Table \ref{tab:reg_results_fg} we present the results of linear regression analysis with fine-grained model types. In this setting, we regard each model type (RoBERTa, ELECTRA, etc.) as a separate case:
\begin{align} \label{eq:regression}
\hat{\nu}_{k}= & \hat{\beta}_{0}+\hat{\beta}_{1}S_{k}+\\ \nonumber
 & +\hat{\beta}_{2}\mathbbm{1}_{T_{k}=1} +\dots+\hat{\beta}_{t}\mathbbm{1}_{T_{k}=t-1} \\ \nonumber
 & +\hat{\beta}_{t+1}S_{k}\cdot \mathbbm{1}_{T_{k}=1} +\dots+\hat{\beta}_{2t}S_{k}\cdot \mathbbm{1}_{T_{k}=t-1} \nonumber
\end{align}
where $S_k$ is the size (in billions of parameters) of model $M_k$ of type $T_k$, and $\mathbbm{1}$ is the indicator function.

As before, whether the model adequately captures the variability of the explained variable (in our case, the variability of $\nu_k$), is measured by $0 \leq R^2 \leq 1$, and large statistically significant coefficients indicate a substantial effect of the corresponding explanatory variables.

To avoid multicollinearity, the estimator of the variance includes $t-1$ model types.  For a model of type 1, the estimated variance is 
$$\hat{\nu}_{k}=(\hat{\beta}_{0}+\hat{\beta}_{2})+(\hat{\beta}_{1}+\hat{\beta}_{t+1})S_{k},$$ whereas for a model of type $t$, it is
$\hat{\nu}_{k}=\hat{\beta}_{0}+\hat{\beta}_{1}S_{k}$. 
Consequently, the effect of the number of parameters in the $t$-th type is incorporated in $\hat{\beta}_{1}$ and serves as the baseline effect, against which all other $t-1$ effects are estimated.

In this case, only Flan-T5 showed a statistically significant dependency between the Size and model type. We note that this was the type with the largest number of models ($=5$), with all others having very few samples ($\leq 3$), and that this analysis suffers from the limitation of small sample sizes for each fine-grained model type.

 \begin{table} 
    \centering
\resizebox{!}{1.15in}{
    \begin{tabular}{lcccc}
         \hline \hline
         & \multicolumn{2}{c}{Wikitext} & \multicolumn{2}{c}{News}\tabularnewline
          & Coeff & P-value & Coeff & P-value\\  
          \hline \hline
Intercept  & 22.59 & 0.728 & 17.93 & 0.712 \tabularnewline
Size & 0.016 & 0.169 & 0.0129 & 0.147 \tabularnewline
Data size & -0.0023 & 0.727 & -0.0018 & 0.713\tabularnewline
T: ELECTRA-g & -18.8194 &	0.77 & -13.8604 & 0.774\tabularnewline
T: RoBERTa & -19.4433 & 0.76 & -15.2669 & 0.749\tabularnewline
T: XLM-R & -18.3511 & 0.77 & -13.6462 & 0.772\tabularnewline
T: Llama 2-C & 1.2206& 0.093 & 0.5091 & 0.3\tabularnewline
T: Flan-T5 & -20.6326 & 0.731 & -15.8104 & 0.725 \tabularnewline
I: S -- ELECTRA-g & -2.3208 & 0.49 & -1.8548 & 0.463\tabularnewline
I: S -- RoBERTa & -1.1458 & 0.723 & -1.0219 & 0.673\tabularnewline
I: S -- XLM-R & -1.7275 & 0.596 & -2.5442& 0.316\tabularnewline
I: S -- Llama 2-C & 0.0017 & 0.908 & -0.0013 & 0.908\tabularnewline
I: S -- Flan-T5 & 0.1473 & \textbf{0.043} & 0.1784 & \textbf{0.007}\tabularnewline
         \hline
    \end{tabular}
    }
    \caption{Analysis of Discrepancy Variance with Model Types. Regression coefficients and corresponding p-values for each model are presented, with MLM models serving as the baseline. P-values below $\alpha=0.05$ are shown in bold.}
    \label{tab:reg_results_fg}
\end{table}

\section{Experiments with Synthetic Data} \label{appendix: B}
\subsection{Data Generation} \label{appendix: B1}

In addition to real datasets, we conducted tests on an automatically generated dataset, which allowed us the ability to control the context and manipulate the occurrence of lower probability tokens.

We used the template: 
\begin{quote}
    [MASK1] [MASK2] is a thing
\end{quote}
and filled in the masks with predetermined spans. 

Noun phrases were extracted from fiction data\footnote{\url{https://huggingface.co/datasets/AlekseyKorshuk/fiction-books}} using SpaCy\footnote{\url{https://spacy.io/}}, and filtered to those that consist of 2 words. Additionally, we used on phrases in which each word is a single token in all the models we tested with. The final dataset consists of $10K$ samples. Sentences in this dataset are not meaningful but are mostly grammatical and resemble real sentences.\footnote{We experimented with additional templates and filling methods; all resulted in qualitatively similar outcomes.}

While we lack expectations about the specific probability values in synthetic data, which can be arbitrarily low, the model is expected to produce consistent probabilities.

\subsection{Results and Analysis} \label{appendix: B2}
\begin{figure}[t]
\centering
\includegraphics[width=0.49\textwidth]{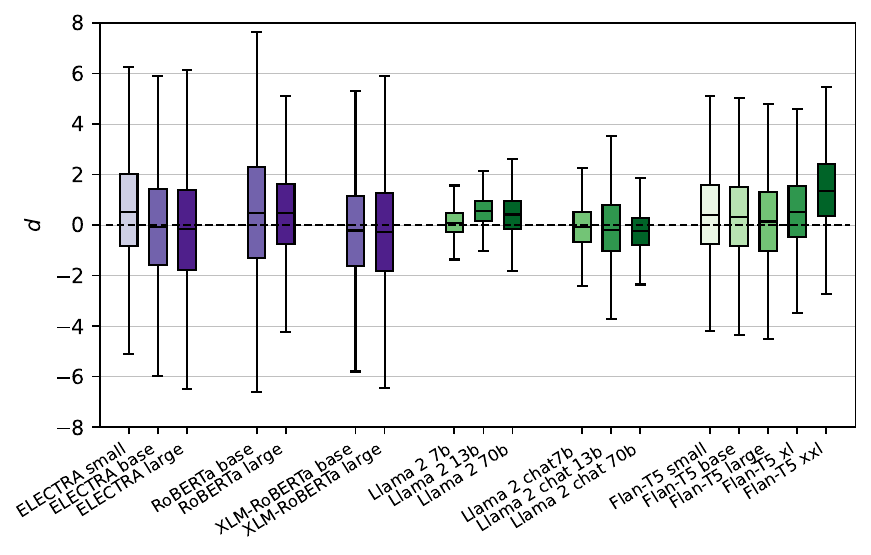}
\caption{Discrepancy Results on the Synthetic datasets.}
\label{fig:synthetic}
\end{figure}

\subsubsection{Consistency}
Here we regard the discrepancy values $d^{(j)}$ for each examined model on the Synthetic dataset (\ref{fig:synthetic}). We can see that, compared to the results for the real datasets (in \ref{fig:subfig_a}, \ref{fig:subfig_b}), the MLM medians are notably further away from 0. The autoregressive models show similar trends for the real and synthetic datasets. 

\paragraph{Discrepancy mean.}
For the Wilcoxon test on the synthetic dataset, all models, except flan-t5-base (adjusted p-value = 0.22), attain a corrected p-value smaller than 0.05.

\paragraph{Discrepancy variance.}
In the Synthetic dataset (\ref{fig:synthetic}), variance is higher compared to real datasets, and no clear trend related to the model size is observed.

\section{Additional Results on Task Comprehension in Autoregressive Models} \label{appendix: C}

\begin{figure}[t]
\centering
\includegraphics[width=0.49\textwidth]{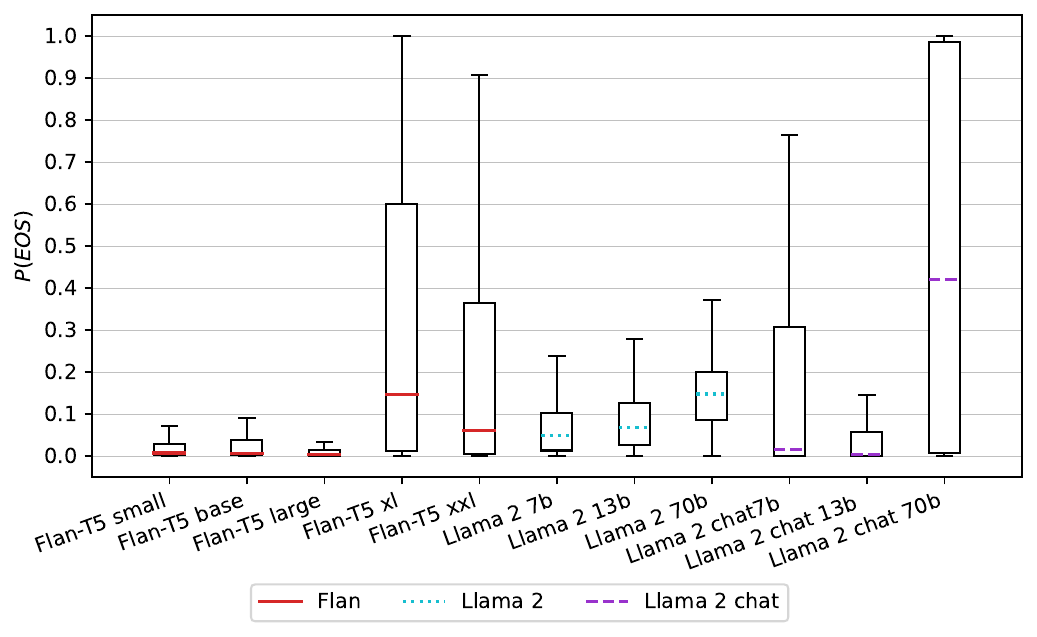}
\caption{
End-of-sentence (EOS) Probabilities for Single-Masked Prediction on the News dataset. Boxes represent the interquartile range (IQR), with whiskers extending up to 1.5 times the IQR. The solid-red, dotted-blue, and dashed-violet lines indicate the medians for Flan-T5, {\sc Llama 2}, and {\sc Llama 2} chat models, respectively. Outliers are omitted for clarity.}
\label{fig:eos}
\end{figure}

In autoregressive models, even when instructed to predict a single word (assuming single-token words), the prediction necessarily consists of at least two tokens: the predicted word and the EOS token. 
. The EOS score provides valuable information, as a probability less than 1 indicates that some probability mass was allocated to spans longer than one token. Therefore, it can be used to assess how well the instruction was ``understood'' by the model, thereby aiding in distinguishing cases where poor predictions arise from a failure to comprehend the task -- a challenge absent in classic MLMs where the infilling task is inherent in the architecture. 

Previous work examines model consistencies through paraphrasing and logical dependencies (see \S \ref{consistency}). However, this type of consistency test faces two primary limitations. First, a model might fail to capture dependencies and similarities between inputs, leading to misinterpretation of imperfect inference capabilities as inconsistencies. Second, inconsistencies in predictions may arise from training data corruption, such as the presence of contradictory facts, instead of model inconsistency. 

In our tests with MLMs, these issues do not apply as the inputs are identical. However, autoregressive models raise concerns regarding the model's ability to fully ``understand'' the instructions. Figure \ref{fig:eos} displays distributions of the EOS probabilities (for the second token) in the News dataset (gathered past the release date).

\begin{figure}[t]
\centering
\includegraphics[width=0.49\textwidth]{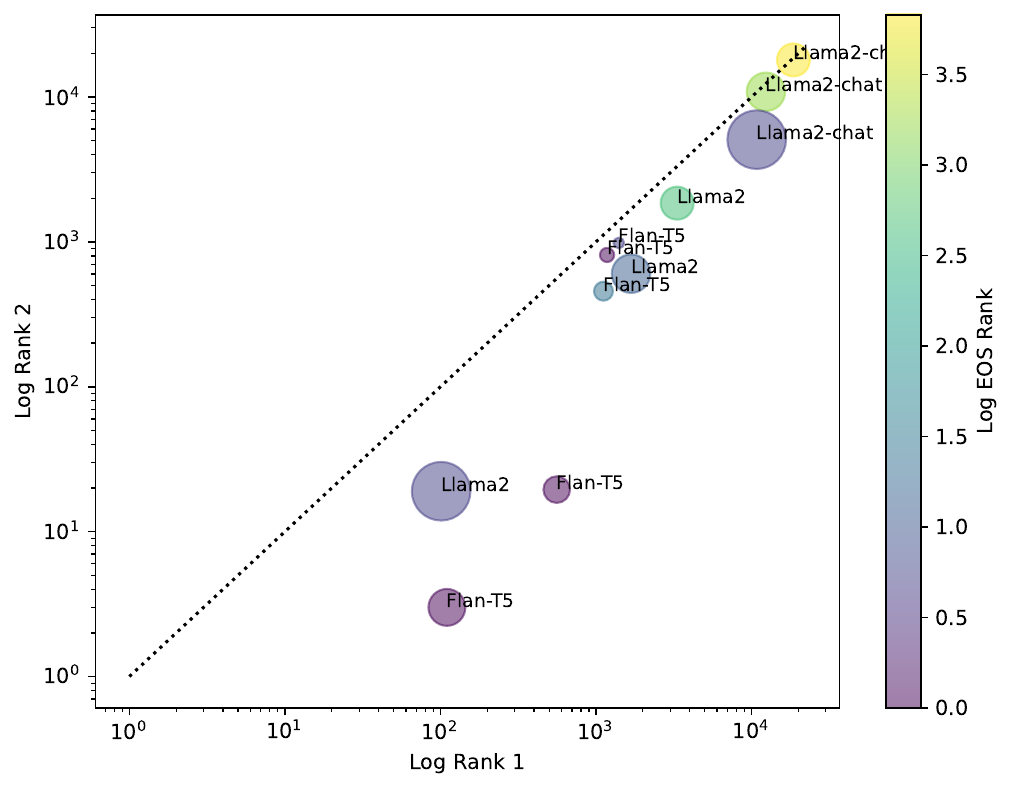}
\caption{Prediction Ranking Summary. The x-axis shows log-scale ranks for single-mask predictions and y-axis displays ranks for double-mask predictions. Each node corresponds to an examined model. Colors indicate EOS prediction ranks (lower ranks indicate better task comprehension), and node size reflects the model's parameter. Results were obtained on a sample of size 200 from the News dataset.}
\label{fig:ranks}
\end{figure}

Figure \ref{fig:ranks} presents an analysis of the assigned ranks for both tokens in autoregressive models, while accounting for model size and EOS scores. The results show alignment between better-performing models and task comprehension, as evidenced by lower ranks for EOS ranks and the second token prediction. An exception is the {\sc Llama 2-Chat} models, which show poor rankings for the real word that cannot be attributed to a lack of task understanding, as several of these models assign low ranks for EOS. Notably, larger models within each model type tend to better task comprehension, in terms of both EOS scores and real word rankings.

\section{Examples for The Effect of Decoding Order} \label{appendix: examples}

To demonstrate the impact of the suggested decoding order on completion quality, we provide examples from news datasets: a case where the decoding order notably affects the outcome, and a case where it does not. Completions are generated using RoBERTa base.

\paragraph{Large Effect.}
To identify cases with large effect, we examined the values of  $\Delta H =  H_{i+1|i} - H_{i|i+1} + H_{i+1} - H_{i}$:  Large $\Delta H$ values suggest that decoding in the suggested order will yield higher probabilities for true tokens, as they correspond to low entropy for double-mask predictions and high entropy for single-mask predictions.

For the following example $ \Delta H =12$ was observed:
\begin{quote}
    with former master's students and co-first <mask> <mask> Pai
\end{quote}
The true masked words are ``authors James''.

The top 10 completions ranked by joint probability, when decoding the first mask first are:

\begin{small}
\begin{enumerate}
\itemsep0em
\item with former master's students and co-first  \textbf{year  student} Pai
\item with former master's students and co-first  \textbf{year  students} Pai
\item with former master's students and co-first  \textbf{president  Ken} Pai
\item with former master's students and co-first  \textbf{president  Michael} Pai
\item with former master's students and co-first  \textbf{chair  Ken} Pai
\item with former master's students and co-first  \textbf{lady  Ken} Pai
\item with former master's students and co-first \textbf{ president  Patrick} Pai
\item with former master's students and co-first  \textbf{president  Fred} Pai
\item with former master's students and co-first  \textbf{secretary  Ken} Pai
\item with former master's students and co-first  \textbf{president  Jeff} Pai
\end{enumerate}
\end{small}

and the top 10 completions when decoding the second mask first are:
\begin{small}
\begin{enumerate}
\itemsep0em
\item with former master's students and co-first \textbf{ president  Ken} Pai
\item with former master's students and co-first\textbf{s  to} Pai
\item with former master's students and co-first  \textbf{author ,} Pai
\item with former master's students and co-first  \textbf{year  Justin} Pai
\item with former master's students and co-first\textbf{s  Chairman} Pai
\item with former master's students and co-first \textbf{ed it} Pai
\item with former master's students and co-first  \textbf{president  chairman} Pai
\item with former master's students and co-first  \textbf{responders  Chairman} Pai
\item with former master's students and co-first  \textbf{year  Michael} Pai
\item with former master's students and co-first \textbf{ did it} Pai
\end{enumerate}
\end{small}

Indeed, in this example, decoding the first mask first yields higher-quality predictions. 
Interestingly, high entropy for the second token often occurs, even when the first token is given, in two-token names. This is expected due to the low predictability of names.

\paragraph{Small Effect.}
Additionally, for comparison, we provide an example of a case where the difference in entropies is small.

For the following example $\Delta H = 1 \cdot 10^{-4}$ was observed:
\begin{quote}
    A group of <mask> <mask> they were well inside the designated safe zone
\end{quote}
The original words are ``friends said''.

The corresponding top 10 completions ranked by joint probability, when decoding the first mask first are:

\begin{small}
\begin{enumerate}
\itemsep0em
\item A group of  \textbf{soldiers  believed} they were well inside the designated safe zone
\item A group of  \textbf{soldiers  signaled} they were well inside the designated safe zone
\item A group of  \textbf{men  believed} they were well inside the designated safe zone
\item A group of  \textbf{soldiers  thought} they were well inside the designated safe zone
\item A group of  \textbf{people  believed} they were well inside the designated safe zone
\item A group of  \textbf{civilians  believed} they were well inside the designated safe zone
\item A group of  \textbf{soldiers  realized} they were well inside the designated safe zone
\item A group of  \textbf{soldiers  ensured} they were well inside the designated safe zone
\item A group of  \textbf{soldiers  felt} they were well inside the designated safe zone
\item A group of  \textbf{men  thought} they were well inside the designated safe zone
\end{enumerate}
\end{small}

and the top 10 completions when decoding the second mask first are:
\begin{small}
\begin{enumerate}
\itemsep0em
\item A group of  \textbf{police  believed} they were well inside the designated safe zone
\item A group of  \textbf{police  thought} they were well inside the designated safe zone
\item A group of  \textbf{protesters  believed} they were well inside the designated safe zone
\item A group of  \textbf{people  thought} they were well inside the designated safe zone
\item A group of  \textbf{people  believed} they were well inside the designated safe zone
\item A group of  \textbf{soldiers  thought} they were well inside the designated safe zone
\item A group of  \textbf{soldiers  believed} they were well inside the designated safe zone
\item A group of  \textbf{protesters  thought} they were well inside the designated safe zone
\item A group of  \textbf{civilians  believed} they were well inside the designated safe zone
\item A group of  \textbf{armed  believed} they were well inside the designated safe zone
\end{enumerate}
\end{small}

In this case, all generations (except for the last one in the second case) yield reasonable completions and therefore no notable differences between the two completion orders were observed.

\end{document}